\documentclass[runningheads]{llncs}
\usepackage[T1]{fontenc}
\usepackage{graphicx}
\usepackage{booktabs}
\usepackage[misc]{ifsym}
\newcommand{\corr}{(\Letter)}
\usepackage{amsmath,amssymb,amsfonts}
\usepackage{microtype}
\usepackage[caption=false]{subfig}
\usepackage[autostyle=true]{csquotes}
\usepackage{multirow}

\usepackage{siunitx}
\usepackage{tikz}
\newcommand*\circled[1]{\tikz[baseline=(char.base)]{
            \node[shape=circle,draw,inner sep=0.6pt] (char) {#1};}}

\usepackage{glossaries}
\glsdisablehyper
\newacronym{AI}{AI}{artificial intelligence}
\newacronym{DNN}{DNN}{deep neural network}
\newacronym{MCU}{MCU}{microcontroller unit}
\newacronym{NAS}{NAS}{neural architecture search}
\newacronym{QAT}{QAT}{quantization-aware training}
\newacronym{PTQ}{PTQ}{post-training static quantization}
\newacronym{MOO}{MOO}{multi-objective optimization}
\newacronym{FLOP}{FLOP}{floating-point operation}
\newacronym{HSS}{HSS}{Hypervolume subset selection}

\usepackage[hidelinks,colorlinks=false]{hyperref}
\usepackage{cite}
\usepackage{color}

\begin{document}

\title{PrototypeNAS: Rapid Design of Deep Neural Networks for Microcontroller Units}

\titlerunning{PrototypeNAS: Rapid Design of DNNs for MCUs}
% If the full title of your paper is short enough to also fit in the running head, you can omit the abbreviated paper title here. You can check as follows: if you comment out the \titlerunning line, something will appear in the header of all odd-numbered pages of your PDF from page 3 onward. This something is either the full title (in which case all is well), or the error message "Title Suppressed Due to Excessive Length". If this error message appears, you're going to want to provide an abbreviated title within the \titlerunning command, because if you won't do it, Springer will do it for you.

%N.B.: Author information (both in the \author{} and \authorrunning{} command) should only be present in the Camera-Ready Version of your paper. The version that you initially submit for review, ought to be double-blind. So, when initially submitting your paper, use:
%\author{Author information scrubbed for double-blind reviewing}
\author{Mark Deutel \corr \and Simon Geis \and Axel Plinge}
% You may leave out the orcidID information, if you want to.
% Use \corr to indicate the corresponding author. Note the spacing around the \corr command. Only one author can be the corresponding author.

%N.B.: comment out the \authorrunning{} command for the double-blind version of your paper submitted for review. Later, if your paper is accepted, use the command for the Camera-Ready Version.
\authorrunning{M. Deutel et al.}
% First names are abbreviated in the running head.
% If there is one author, write 'A.L. Benjamin'.
% If there are two authors, write 'A.L. Benjamin and C.C. Broadus Jr.'
% If there are more than two authors, '[...] et al.' is used.

\institute{Fraunhofer Institute for Integrated Circuits, Fraunhofer IIS, Germany\\\email{\{mark.deutel, simon.geis, axel.plinge\}@iis.fraunhofer.de}}
%\institute{Fictional Southern University, Savannah GA 31404, USA \email{\{a.l.benjamin,a.a.patton\}@fsu.fake}
%\and
%Fictional West Coast University, Long Beach CA 90840, USA \email{ccb@fwcu.fake}
%\and
%Secondary European Affiliation, Tiergartenstr. 17, 69121 Heidelberg, Germany
%\email{lncs@springer.com}}

\toctitle{PrototypeNAS: Rapid Design of Deep Neural Networks for Microcontroller Units}
\tocauthor{Mark~Deutel, Simon~Geis, Axel~Plinge}

\maketitle              % typeset the header of the contribution

\begin{abstract}
Enabling efficient \gls{DNN} inference on edge devices with different hardware constraints is a challenging task that typically requires \gls{DNN} architectures to be specialized for each device separately.
To avoid the huge manual effort, one can use \gls{NAS}. However, many existing \gls{NAS} methods are resource-intensive and time-consuming because they require the training of many different \glspl{DNN} from scratch. Furthermore, they do not take the resource constraints of the target system into account.
To address these shortcomings, we propose PrototypeNAS, a zero-shot \gls{NAS} method to accelerate and automate the selection, compression, and specialization of \glspl{DNN} to different target \glspl{MCU}.
We propose a novel three-step search method that decouples \gls{DNN} design and specialization from \gls{DNN} training for a given target platform. First, we present a novel search space that not only cuts out smaller \glspl{DNN} from a single large architecture, but instead combines the structural optimization of multiple architecture types, as well as optimization of their pruning and quantization configurations. Second, we explore the use of an ensemble of zero-shot proxies during optimization instead of a single one. Third, we propose the use of \gls{HSS} to distill \gls{DNN} architectures from the Pareto front of the \gls{MOO} that represent the most meaningful tradeoffs between accuracy and \glspl{FLOP}. 
We evaluate the effectiveness of PrototypeNAS on 12 different datasets in three different tasks: image classification, time series classification, and object detection. Our results demonstrate that PrototypeNAS is able to identify \glspl{DNN} within minutes that are small enough to be deployed on off-the-shelf \glspl{MCU} and still achieve accuracies comparable to the performance of large \gls{DNN} architectures.
\end{abstract}
\glsresetall
\section{Introduction}
\label{ref:introduction}

We address the problem of quickly and efficiently designing and training \glspl{DNN} for inference on resource-constrained \glspl{MCU}. \Glspl{DNN} have become the de facto standard for data analysis and machine learning tasks. However, with the growth of \glspl{DNN} in both size and computational cost over the last years, running \glspl{DNN}, especially on a diverse set of different resource constrained embedded systems and at low latency, has become a major challenge.

To address this problem, we propose PrototypeNAS, a zero-shot \gls{NAS} framework for rapidly designing \glspl{DNN} for \glspl{MCU} with different hardware and resource constraints. Compared to other hardware-aware \gls{NAS} frameworks, PrototypeNAS is novel in three aspects. First, it uses an ensemble of zero-shot proxies that compete as objectives in a \gls{MOO} rather than being weighted and linearized into a single ensemble proxy score. Second, it implements a novel search space that combines architecture selection, size and structure optimization, and optimization of pruning and quantization configuration, instead of optimizing them as unrelated problems. Third, it introduces \gls{HSS} to further refine the Pareto optimal models from optimization to a set of 3-5 models that cover the most meaningful tradeoffs between accuracy and resource consumption.

Additionally, in our extensive evaluation, we demonstrate the effectiveness and versatility of PrototypeNAS by using it to find optimized \glspl{DNN} deployable on an ARM Cortex-M \gls{MCU} for 12 datasets from three tasks: image classification, time series classification, and object detection. We also compare PrototypeNAS to two other hardware-aware NAS methods, TinyNAS (MCUNet)~\cite{lin2020mcunet} and NATS-Bench~\cite{dong2021nats}. On average, the \gls{DNN} architectures found by PrototypeNAS outperform the ones proposed by the other two frameworks by 5\,\% in accuracy on the CIFAR10 dataset.

As a result, PrototypeNAS is a resource- and time-efficient way to search for \gls{DNN} architectures without having to train hundreds of \gls{DNN} architectures to find a single good candidate. Instead, PrototypeNAS reduces the training to only 3-5 \gls{DNN} candidates while still considering hundreds of architectures in its search.

\section{Related Work}
\label{ref:related}

\Gls{NAS} is a set of techniques used to automate the \gls{DNN} architecture design process, generally by solving an optimization problem. While the initial focus was on maximizing accuracy and using reinforcement learning to control the search process~\cite{zoph2018learning}, more recently, hardware-aware \gls{NAS}~\cite{tan2019mnasnet}, i.e., considering memory and inference speed in addition to accuracy as a \gls{MOO}, has become a major focus of research. A subset of this work focuses more specifically on \gls{NAS} for resource-constrained devices such as \glspl{MCU}~\cite{cai2020once,lin2020mcunet,heidorn2026entropy}. 

In addition, zero-shot \gls{NAS}~\cite{lee2018snip,mellor2021neural,jiang2023meco,li2023zico} has received considerable attention to address the problem of having to train a large number of \gls{DNN} candidates, for example, when using black-box optimization~\cite{deutel2025combining} or reinforcement learning~\cite{zoph2018learning}. All zero-shot \gls{NAS} techniques rely on using a proxy metric for accuracy to avoid training. As a result, a large number of different zero-shot proxies have been proposed, focusing on different features that can be computed from an untrained \gls{DNN}, such as the Pearson correlation matrix of the intermediate feature maps~\cite{jiang2023meco} or the number of linear regions in the input space~\cite{mellor2021neural}. In addition, zero-shot \gls{NAS} techniques, especially when used for hardware-aware \gls{NAS}, often propose the use of a large pre-trained \enquote{super-net} from which smaller networks more suited to the constraints of the target hardware are then derived and tuned~\cite{cai2020once}.

While zero-shot \gls{NAS} works to find a sufficiently good architecture in a relatively short time, there are still challenges to overcome. Zero-shot proxies are known to be imprecise and bias certain \gls{DNN} architectures~\cite{jing2025zero,bhardwaj2023zico}, often leading to poor correlation with actual training accuracy. Furthermore, the reliance on a single large supernet means that the resulting search space is limited to exploitation around a pre-existing optimum and never explores the full architectural landscape of \glspl{DNN}.

In parallel to \gls{NAS}, other techniques for \gls{DNN} compression have been explored. A significant number of efficient human-engineered \gls{DNN} architectures have been proposed over the years. Starting with SqueezeNet~\cite{iandola2016squeezenet}, MobileNet~\cite{sandler2018mobilenetv2}, and EfficientNet~\cite{tan2019efficientnet}, to more recent designs such as MCUNet~\cite{lin2020mcunet}, which specifically targets microcontrollers, and ConvNeXt~\cite{liu2022convnet}, which combines a traditional CNN with transformer elements. In addition, pruning and quantization have become common techniques for designing efficient \gls{DNN} architectures for resource-constrained \glspl{MCU}~\cite{deutel2023energy}.
\begin{figure}[t]
    \centering
    \includegraphics[width=\linewidth]{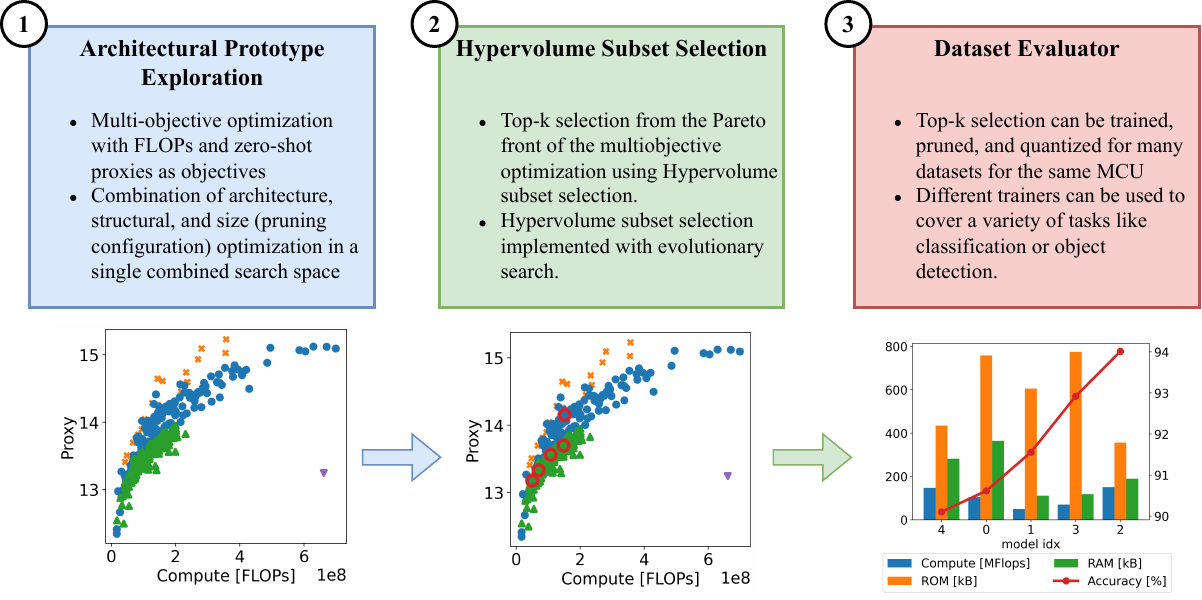}
    \caption{Schematic of the three step pipeline of PrototypeNAS.}
    \label{fig:overview}
\end{figure}

\section{Method}
\label{sec:method}

PrototypeNAS is a zero-shot \gls{NAS} method that enables rapid exploration of optimized \glspl{DNN} for \gls{MCU} deployment by decoupling design space exploration, i.e., the search for prototypical network architectures and their compression configuration, from training on the target datasets.

We give a schematic overview of the three-step method of PrototypeNAS in Fig.~\ref{fig:overview}. \circled{1}~First, a constrained multi-objective optimization is performed with the ensemble of zero-shot proxies and the number of \glspl{FLOP} as a proxy for the computational cost of \gls{DNN} inference as objectives. The optimization is constrained by the memory limits of the target \gls{MCU} to ensure that the exploration focuses only on \gls{DNN} architectures that can actually be deployed on the target. The search space of the optimization combines both architectural selection from a pool of predefined baseline architectures and their structural and size optimization, i.e., pruning parameter configuration, into a single search space. \circled{2}~Second, a top-$k$ selection is performed from the resulting Pareto set using \gls{HSS}. \circled{3}~Finally, the resulting selection of optimized \gls{DNN} architectures is trained, pruned, and quantized on the target datasets. In the following sections, we describe each of the three steps of PrototypeNAS in detail.

\begin{table}[t]
    \centering
    \caption{Search space $X$ of the \gls{MOO}, $\forall i \in \{0, 1, 2, 3\}$ resulting in 14 tunable hyperparameters in total.}
    \label{tab:method:searchspace}
    \begin{tabular*}{\textwidth}{@{\extracolsep{\fill}}llcc}
        \toprule
        \textbf{Optimization} & \textbf{Hyperparameter} & \textbf{Type} & \textbf{Range/Values} \\
        \midrule
        architecture & baseline architecture & categorical & task dependent \\
        \midrule
        \multirow{2}{*}{structural} & group depth $i$ & categorical & $\left[0, 1, 2, 3\right]$ \\
        & kernel\,\&\,stride $i$ & categorical & $\left[\left[3, 2\right], \left[3, 1\right], \left[5, 2\right], \left[5, 1\right], \left[7, 2\right], \left[7, 1\right]\right]$ \\
        \midrule
        \multirow{ 2}{*}{size} & width multiplier & continuous & $\left[0.1, 1.0\right]$ \\
        & pruning sparsity $i$ & continuous & $\left[0.1, 0.9\right]$ \\
        \bottomrule
    \end{tabular*}
\end{table}

\subsection{Architectural Prototype Exploration}
\label{subsec:method:exploration}

We formulate the architectural prototype exploration as a constrained \gls{MOO} in Eq.~\eqref{eq:moopt}. The objectives are to minimize the number of \glspl{FLOP} of a model $flops(x)$ while maximizing an ensemble of four proxy metrics $prox_{i}(x)$ with $i \in \{0,\dots,3\}$. The four proxies used in this work are MeCo~\cite{jiang2023meco}, ZiCo~\cite{li2023zico}, NASWOT~\cite{mellor2021neural}, and SNIP~\cite{lee2018snip}, which we selected due to the different \gls{DNN} features they use for evaluation, compare~\cite{huang2025evolving}. The constraints of the optimization $ram_{max}$, $rom_{max}$, and $flops_{max}$ are derived from the hardware limits of the targeted MCU.

\begin{equation}
\begin{aligned}\label{eq:moopt}
    \min_{x \in X} \quad &flops(x), -prox_{1}(x), \dots, -prox_{3}(x) \\
    \textrm{s.t.} \quad ram(x) &\leq ram_{max} \\
    rom(x) &\leq rom_{max} \\
    flops(x) &\leq flops_{max}
\end{aligned}
\end{equation}

We designed the search space from which PrototypeNAS samples \glspl{DNN}, focusing on emulating the process that a specialist makes when designing an efficient \gls{DNN} architecture for an \gls{MCU}. Based on this premise, we make the following assumptions: First, there is a set of baseline architectures from which \gls{DNN} candidates can be derived. Second, each of the baseline architectures can be abstracted as follows: An initial layer pattern, followed by a set of repeatable layer patterns (\textit{superblocks}), followed by a classifier. Third, a pruning and quantization scheme is defined that is executed during training to dynamically compress and scale down the model during training. Based on these three assumptions, we define the search space $X$ in Table~\ref{tab:method:searchspace}.

\begin{figure}[t]
    \centering
    \includegraphics[width=\linewidth]{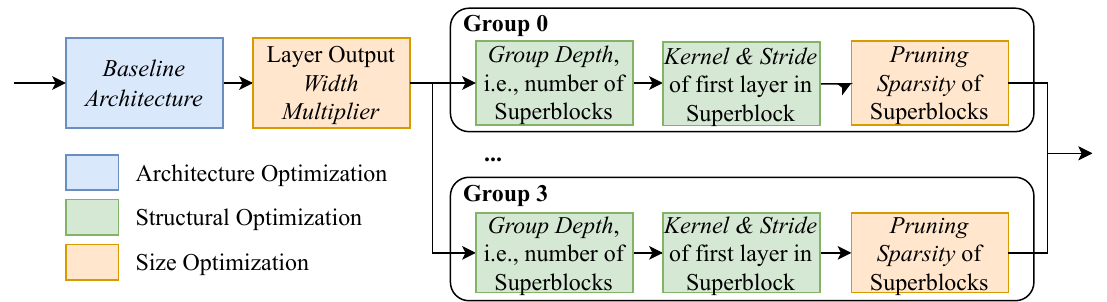}
    \caption{The proposed search space and how a \gls{DNN} prototype can be created from it. Each baseline architecture consists of repeatable superblocks, i.e., a predefined pattern of layers, organized in the search space into four groups to be optimized separately.}
    \label{fig:searchspace}
\end{figure}

In the following, we describe how $X$ is structured and how \gls{DNN} prototypes can be created from it, see Fig.~\ref{fig:searchspace} for a schematic overview. The \textit{architecture} hyperparameter allows the optimizer to choose a baseline \gls{DNN} architecture from a pool of predefined architectures. In the scope of this work, six \gls{DNN} architectures are supported for image classification, two architectures for time series classification, and one for object detection (see Section~\ref{sec:evaluation} for details).

The backbone of the selected baseline \gls{DNN} is then split into repeatable superblocks as described above, for example depthwise separable convolutions in MobileNetV2. How each architecture is split into superblocks is defined prior. The superblocks are then organized into four groups. Each group contains at least one super block and up to four additional blocks, which is controlled for each group via the \textit{group depth} hyperparameter. In addition, each group has an optimizable \textit{kernel}\,\&\,\textit{stride} hyperparameter which is used to configure the first convolution of each superblock of the group. Together, these hyperparameters allow structural optimization of the baseline \glspl{DNN}.

The part of the search space allowing for size optimization of the baseline \glspl{DNN} consists of the \textit{width multiplier} and \textit{pruning sparsity} hyperparameters. The \textit{width multiplier} parameter controlls the initial size of the baseline \gls{DNN} architecture at the beginning of training by scaling the output channels in all convolutional layers uniformly. In addition, an iterative pruning schedule is defined that will be executed during training. For each of the four groups of superblocks, a separate \textit{pruning sparsity} hyperparameter controls the target percentage of channels to be removed by pruning.

The objective function evaluation for a set of hyperparameters proposed by the optimizer consists of three steps: First, the selected baseline architecture is initialized and configured according to the structural optimization. Second, pruning and quantization are applied (without any training) to query parameter and size reduction. Third, the resulting model is translated to C code (cf.~\cite{deutel2023energy}) to get the actual ROM and RAM usage when deployed on the targeted MCU. To calculate the number of \glspl{FLOP} of a \gls{DNN}, Pytorch's built-in profiler is used. Since the objective function evaluation does not require any training due to the use of zero-shot proxies, sampling is both resource and time efficient, allowing for rapid exploration of the search space.

\subsection{Hypervolume Subset Selection}
\label{subsec:method:hss}

We obtain a set of Pareto-optimal architecture and compression configurations from the prototype exploration described in the previous section. The size of this set can be between one and as many solutions as there were \glspl{DNN} explored during the optimization. However, in practice, a set of 3-5 architectural tradeoffs is usually sufficient for a decision maker to select a \gls{DNN} for a given use case. A smallest model, a model with the highest accuracy, and 1-3 tradeoffs in between. To avoid situations where large Pareto sets proposed by the optimization have to be trained, compressed, evaluated, and finally presented to the decision maker, we implemented a \gls{HSS} algorithm based on evolutionary programming and inspired by algorithms like~\cite{bader2011hype} that use Hypervolume as a selection criterion.

Let $H(A)$ be the Hypervolume indicator of a set of solutions $A$ as described in~\cite{zitzler2002multiobjective}. We formulate an optimization problem in Eq.~\eqref{eq:hss} which aims to find a set $A \subset P$ with $|A| = k$ and $k \in \{1, 2, \dots, |P|\}$ that maximizes $H(A)$ where $P$ is the full Pareto set resulting from the prototype exploration.

\begin{equation}
    \label{eq:hss}
    \max_{\substack{A \subset P\\|A| = k}} H(A)
\end{equation}

We solve this optimization problem using a greedy selection strategy that retains solutions with higher Hypervolume contributions. First, we encode $P$ as a binary gene $g$, where each bit in $g$ corresponds to a solution in $P$, which is either set to $1$ if the sample should be part of $A$, or $0$ if not. Consequently, for each gene, only $k$ bits can be set to $1$, while all other bits are set to $0$. We then evolve an initial population $G \in \{g_0, g_1, \dots, g_{n-1}\}$ of a given size $n$, using mutation and crossover operators, and the Hypervolume indicator to evaluate the fitness of each $g \in G$.

Naturally, crossover, mutation, and the initial generation of $G$ can lead to the creation of \enquote{invalid} genes, i.e. genes with more than $k$ bits set to $1$. Our algorithm greedily repairs such genes. If there are less than $k$ bits set to $1$, it iteratively sets bits to $1$ until exactly $k$ bits are set to $1$. This has no negative effect on the fitness of the gene, since the Hypervolume can only increase or stay the same with additional solutions added, but never decrease. If there are more than $k$ bits set to $1$, the algorithm instead tries to set additional bits to zero while keeping the Hypervolume as high as possible. To do this, it sets every bit that is $1$ to $0$ one by one and recalculates the Hypervolume. Then the algorithm sorts all bits by their calculated Hypervolumes in ascending order. Finally, it selects the top $k$ bits whose removal has the largest negative impact on the fitness of the gene, while setting all other bits to $0$.

In all our experiments, we configured the Hypervolume subset selection algorithm with an initial population size of $2000$, a mutation rate of $0.3$, and ran it for $10\,000$ generations.

\subsection{Dataset Evaluator}
\label{subsec:method:dataset}

To train, prune, and quantize the subsets of \glspl{DNN} designed by PrototypeNAS, we implement two dataset evaluators. One evaluator is for training image and time series classification tasks using Pytorch Lightning, while the other one is for training the object detection task with the YOLOv5 framework. As baseline architectures for PrototypeNAS we used implementations provided by torchvision and torchaudio with slight modifications so that they can be generated from a set of hyperparameters from the search space introduced in Section~\ref{subsec:method:exploration}. 

We also implemented iterative structure pruning and quantization for both trainers. We used \gls{PTQ} for the image classification and object detection tasks, and \gls{QAT} with 15 additional training epochs for the time series classification task. Since \gls{PTQ} is applied to \glspl{DNN} after training, it is generally faster to use and less computationally intensive, while \gls{QAT} is more resource intensive since it is applied during training, but also more accurate. While we found during our experiments that \gls{PTQ} worked reliably well for the two image-based tasks, we experienced significant drops in accuracy for the time series datasets when using \gls{PTQ}. As a result, we used \gls{QAT} for the time series datasets, as it yielded significantly better accuracies. This is most likely due to \gls{QAT}'s ability to better account for the varying activation ranges resulting from the time series input.
\section{Evaluation}
\label{sec:evaluation}

We evaluate PrototypeNAS on three different tasks and 12 datasets in total: image classification for the CIFAR10~\cite{krizhevsky2009learning}, CIFAR100~\cite{krizhevsky2009learning}, GTSRB~\cite{houben2013detection}, Flowers~\cite{nilsback2008automated}, Birds~\cite{wah2011birds}, Cars~\cite{krause20133d}, Pets~\cite{parkhi2012cats}, and ArxPhotos314~\cite{arxnetsa2025arxphoto} datasets, time series classification for the Daliac~\cite{leutheuser2013hierarchical}, MAFULDA\footnote{\url{https://www02.smt.ufrj.br/~offshore/mfs/page_01.html}}, and BitBrain Sleep~\cite{larraz2025bitbrain} datasets, and person detection using a subset of COCO~\cite{lin2014microsoft}.

For the three tasks, we first performed the prototype exploration as described in Section~\ref{subsec:method:exploration} for $500$ trials, then performed the \gls{HSS} as described in Section~\ref{subsec:method:hss}, and finally evaluated the top-$5$ selection of each of the tasks on the respective datasets. As constraints of the optimization, we used $ram_{max}=\SI{480}{KB}$, $rom_{max}=\SI{2}{MB}$, and $flops_{max}=\si{1e9}$ to describe the resource constraints of the iMXRT1062, a Cortex-M7 \gls{MCU}, which we selected as the deployment target. For training, we used the same configuration for all datasets: A batch size of $48$, stochastic gradient descent with a learning rate of $0.001$ and a momentum of $0.9$, and $100$ training epochs.

We provide a detailed insight and discussion of our results in Section~\ref{subsec:evaluation:classification}~and~\ref{subsec:evaluation:detection}. Furthermore, in Section~\ref{subsec:evaluation:proxy}, we discuss the precision of the proxy ensemble used in PrototypeNAS during exploration. In Section~\ref{subsec:evaluation:comparison}, we give a comparison of PrototypeNAS to two other \gls{NAS} frameworks, TinyNAS (MCUNet)~\cite{lin2020mcunet} and NATS-Bench~\cite{dong2021nats}. Finally, in Section~\ref{subsec:evaluation:ablation} we measure the impact of PrototypeNAS on performance and emissions of the \gls{NAS} algorithm.

\begin{figure}[t]
    \centering
    \subfloat[Image classification\label{fig:class:optim:vision}]{%
        \includegraphics[width=\textwidth]{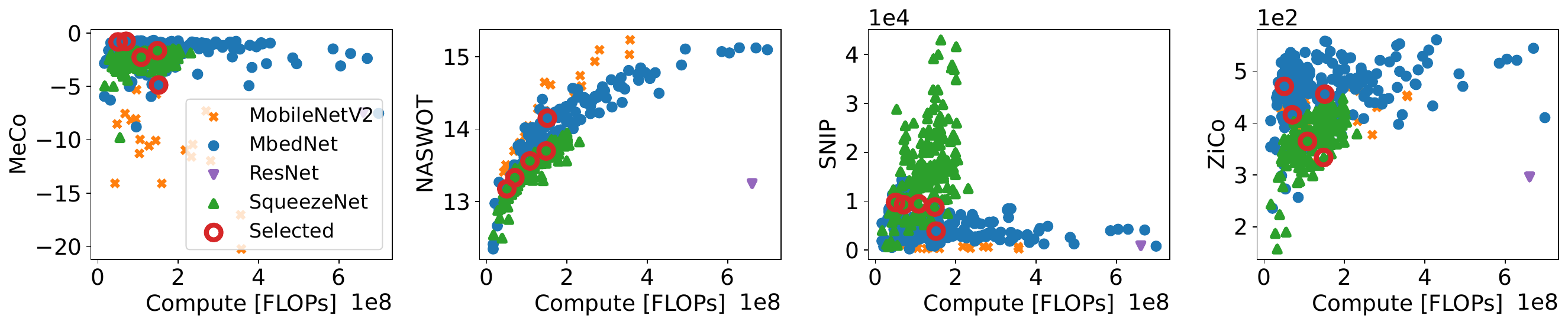}%
    }\\
    \subfloat[Time series classification\label{fig:class:optim:time}]{%
        \includegraphics[width=\textwidth]{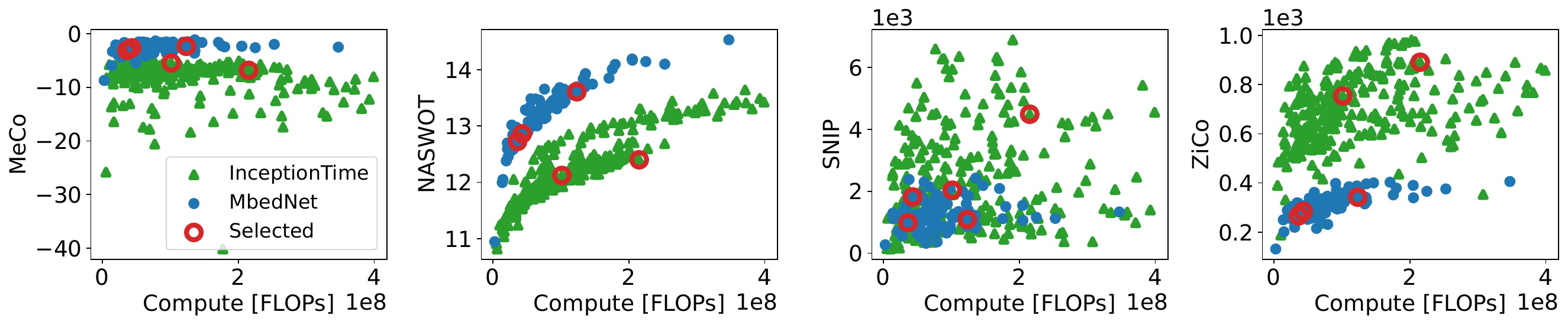}%
    }
    \caption{Optimization results. The top-5 \glspl{DNN} selected from the Pareto front by \gls{HSS} are marked with red circles.}
    \label{fig:class:optim}
\end{figure}

\subsection{Results for Image and Time Series Classification}
\label{subsec:evaluation:classification}

Fig.~\ref{fig:class:optim} shows the results of the prototype exploration for image and time series classification and the four proxy scores MeCo, NASWOT, SNIP, and ZiCo\footnote{\glspl{DNN} and pre-trained weights: \url{https://doi.org/10.5281/zenodo.18878249}}. We performed separate optimizations for the image and time series datasets because we used different sets of baseline architectures for the two tasks, four \gls{DNN} architectures for image classification and two for time-series classification, which we denote in the plots with different colors and markers. For image classification, our baseline set includes MobileNetV2~\cite{sandler2018mobilenetv2}, ResNet18~\cite{he2016deep}, Squeezenet~\cite{iandola2016squeezenet}, and MBedNet (our own \gls{DNN} architecture derived from MobileNet), see Fig.~\ref{fig:class:optim:vision}. For time series classification, we used InceptionTime~\cite{ismail2020inceptiontime} and a version of MBedNet where we replaced all 2D convolutions with their 1D counterparts, see Fig.~\ref{fig:class:optim:time}. We used a $128 \times 128$ pixel input during all experiments for image classification, while for the time series classification experiments we split the input into windows of constant length and used them without any further preprocessing. After optimization, we identified five models from the final Pareto front using \gls{HSS} as described in Section~\ref{subsec:method:hss} and marked the selected models in Fig.~\ref{fig:class:optim} with red circles.

\begin{table}[t]
    \centering
    \caption{Test accuracy of the trained, pruned, and quantized \glspl{DNN} found by PrototypeNAS for the image datasets. Latency and energy was measured on an iMXRT1062 Cortex-M7 \gls{MCU} using an external Joulescope energy analyzer.}
    \label{tab:results:vision}
    \begin{tabular*}{\textwidth}{@{\extracolsep{\fill}}l*{5}{S[table-format=3.1,table-model-setup=\bfseries,text-series-to-math]}}
        \toprule
        \textbf{Optim. Index} & {210} & {283} & {190} & {311} & {237} \\
        \midrule
        \textbf{Architecture} & {MBedNet} & {MBedNet} & {SqueezeNet} & {SqueezeNet} & {MBedNet} \\
        \midrule
        \multicolumn{6}{l}{\textbf{Quantized Test Accuracy [\%] (\gls{PTQ})}} \\
        CIFAR10 & 91.8 & 92.5 & 90.3 & 90.0 & \bfseries 93.7 \\
        CIFAR100 & 67.4 & 70.5 & 66.5 & 63.7 & \bfseries 72.3 \\
        ArxPhotos314 & 90.8 & 95.4 & \bfseries 97.4 & 79.5 & 93.6 \\
        Flowers & 88.9 & 91.6 & 82.9 & 86.3 & \bfseries 93.3 \\
        Birds & 62.8 & 65.7 & 59.5 & 55.6 & \bfseries 69.9 \\
        Pets & 99.8 & 99.8 & 97.5 & 96.1 & \bfseries 99.9 \\
        Cars & 64.8 & 64.1 & 59.8 & 55.2 & \bfseries 76.4 \\
        GTSRB & 96.3 & 96.1 & 95.8 & \bfseries 97.0 & 95.5 \\
        \midrule
        Compute [MFLOPs] & 50.1 & 70.0 & 105.0 & 147.5 & 151.0 \\
        ROM [KB] & 605.4 & 774.9 & 758.4 & 434.7 & 356.5 \\
        RAM [KB] & 111.7 & 118.3 & 364.7 & 281.6 & 189.7 \\
        \midrule
        Latency [ms] & \multicolumn{1}{S[table-format=3.1(1.1)]}{224.1(0.1)}{} & \multicolumn{1}{S[table-format=3.1(1.1)]}{312.9(0.1)}{} & \multicolumn{1}{S[table-format=3.1(2.1)]}{367.1(0.1)}{} & \multicolumn{1}{S[table-format=3.1(2.1)]}{447.2(0.1)}{} & \multicolumn{1}{S[table-format=3.1(2.1)]}{753.2(0.1)}{} \\
        Energy [mJ] & \multicolumn{1}{S[table-format=3.1(1.1)]}{77.6(8.6)}{} & \multicolumn{1}{S[table-format=3.1(1.1)]}{108.6(9.9)}{} & \multicolumn{1}{S[table-format=3.1(2.1)]}{126.0(12.7)}{} & \multicolumn{1}{S[table-format=3.1(2.1)]}{156.3(16.8)}{} & \multicolumn{1}{S[table-format=3.1(2.1)]}{254.5(13.7)}{} \\
        \bottomrule
    \end{tabular*}
\end{table}

For both image and time series optimization, the results in Fig.~\ref{fig:class:optim} show that the four proxies rank the base architectures of their respective search spaces differently. For example, SNIP gave a significantly higher score to almost all architectural variants of SqueezeNet than any of the other three proxies in Fig.~\ref{fig:class:optim:vision}. Another example of this apparent \enquote{disagreement} among the proxies can be seen in Fig.~\ref{fig:class:optim:time}, where NASWOT and MeCo rank MBedNet higher than InceptionTime, while the other two proxies do the opposite.

Another observation that can be made is that all \glspl{DNN} of a single architecture type cluster around their unmodified baseline architecture in the target space. This is not unexpected, since once a baseline architecture is selected, the other hyperparameters in PrototypeNAS's search space focus on modifying the structure of the baseline architecture within the limits of its predefined superblocks, or scaling the architecture in width, but do not fundamentally change its design. This is intentional, as the goal of PrototypeNAS is to quickly fine-tune a \gls{DNN} to fit the constraints of a given \gls{MCU} and not to come up with completely new \gls{DNN} designs from scratch. Therefore, another way to interpret the results in Fig.~\ref{fig:class:optim} is that each baseline architecture represents a local optimum, with the search space being designed in a way to encourage an efficient search around it.

Regarding the difference in proxy scoring observed in our results, we point to the general consensus found in related work that zero-shot proxies are often imprecise and tend to favor certain architectures~\cite{jing2025zero,bhardwaj2023zico}. As a result, and similar to related work such as \cite{huang2025evolving}, this motivates us to consider the evaluation of multiple proxies to guide our optimization, rather than relying on a single proxy. In contrast, we do not attempt to weight and linearize the proxies to form a new score, but instead treat them as competing objectives in our \gls{MOO}.

\begin{table}[t]
    \centering
    \caption{Test accuracy of the trained, pruned, and quantized \glspl{DNN} found by PrototypeNAS for the time series datasets. Latency and energy was measured on an iMXRT1062 Cortex-M7 \gls{MCU} using an external Joulescope energy analyzer.}
    \label{tab:results:time}
    \begin{tabular*}{\textwidth}{@{\extracolsep{\fill}}l*{5}{S[table-format=3.1,table-model-setup=\bfseries,text-series-to-math]}}
        \toprule
        \textbf{Optim. Index} & {224} & {226} & {73} & {273} & {347} \\
        \midrule
        \textbf{Architecture} & {MBedNet} & {MBedNet} & {Inception} & {MBedNet} & {Inception} \\
        \midrule
        \multicolumn{6}{l}{\textbf{Quantized Test Accuracy [\%] (\gls{QAT})}} \\
        BitBrain & 83.7 & 87.2 & 81.7 & \bfseries 88.0 & 78.9 \\
        Mafaulda & 96.8 & 98.3 & 97.4 & 97.8 & \bfseries 98.5 \\
        Daliac & 95.9 & 97.1 & \bfseries 97.2 & 97.0 & 96.1 \\
        \midrule
        Compute [MFLOPs] & 36.3 & 43.2 & {101.5} & {123.5} & {215.3} \\
        ROM [KB] & 149.0 & 231.0 & 571.0 & 337.0 & 978.0 \\
        RAM [KB] & 246.0 & 244.0 & 251.0 & 252.0 & 250.0 \\
        \midrule
        Latency [ms] & \multicolumn{1}{S[table-format=3.1(1.1)]}{162.3(0.1)} & \multicolumn{1}{S[table-format=3.1(1.1)]}{183.6(0.1)} & \multicolumn{1}{S[table-format=3.1(2.1)]}{447.2(0.1)} & \multicolumn{1}{S[table-format=3.1(2.1)]}{467.9(0.3)} & \multicolumn{1}{S[table-format=3.1(2.1)]}{635.0(1.4)} \\
        Energy [mJ] & \multicolumn{1}{S[table-format=3.1(1.1)]}{55.8(5.7)} & \multicolumn{1}{S[table-format=3.1(1.1)]}{63.7(6.4)} & \multicolumn{1}{S[table-format=3.1(2.1)]}{156.0(16.7)} & \multicolumn{1}{S[table-format=3.1(2.1)]}{165.8(15.9)} & \multicolumn{1}{S[table-format=3.1(2.1)]}{227.6(17.6)} \\
        \bottomrule
    \end{tabular*}
\end{table}

As a result, the top $k=5$ architectures selected by the \gls{HSS} at the end of the optimization do not directly follow the ranking of any of the individual proxies and \glspl{FLOP}, but are instead derived from the Pareto front between \glspl{FLOP} and all four proxies. As a result, PrototypeNAS avoids being biased toward particular architectures and achieves a balanced evaluation.  

We show the test accuracy on all evaluated datasets of the five trained, pruned, and quantized DNN prototypes selected by \gls{HSS} in Table~\ref{tab:results:vision} (image classification) and Table~\ref{tab:results:time} (time series classification) sorted in ascending order by \glspl{FLOP}. For the image classification models we used \gls{PTQ} for quantization, while for the time series models we used \gls{QAT}. Furthermore, for each of the five DNNs shown in the two tables, we report the base architecture from which the prototype was derived, the model's RAM and ROM requirements in kilobytes, reflecting the actual memory requirements on the target MCU, as well as the average latency and energy consumption of the \glspl{DNN} measured externally using a Joulescope energy analyzer when executed on an iMXRT1062 Cortex-M7 MCU. For each dataset, we highlighted the model with the highest test accuracy. For time series classification, we trained with randomly initialized weights, while for image classification, we performed 50 epochs of ImageNet pre-training.

For both the image and time series classification tasks, the selection of \glspl{DNN} found by PrototypeNAS achieved accuracies competitive with or better than related embedded \gls{DNN} architectures, see also Section~\ref{subsec:evaluation:comparison} for a direct comparison of our approach with other \gls{NAS} methods. Furthermore, while a correlation between \glspl{FLOP} and accuracy can be observed within an architecture type, e.g. both the MBedNet-based architectures 210, 283, and 237 and the SqueezeNet-based architectures 190 and 311 in Table~\ref{tab:results:vision} show a linear correlation between \glspl{FLOP} and accuracy, this correlation cannot be observed when ranking across the two different architecture types. Since the PrototypeNAS search space optimizes multiple architecture types together, ranking by \glspl{FLOP} is not sufficient, meaning that the structure and expressiveness of the architectures must also be considered, motivating our use of zero-shot proxies in this work. 

In addition, when considering resource consumption, we noticed that the RAM requirements are very similar across all five \glspl{DNN} for both the image and time series classification tasks, even though they have very different compute and ROM requirements. The reason is that the deployment framework we utilize for our experiments reuses memory for multiple intermediate feature maps during inference. This means that the larger initial feature maps typically dominate the overall RAM requirements, since they are often similar in size, as all models share the same input size. As a result, we are only modeling memory requirements as constraints rather than objectives during optimization. Finally, our results show that latency and energy per sample scale linearly with \glspl{FLOP}, making it a good proxy for optimizing these two metrics without having hardware in the loop.

\begin{table}[t]
    \centering
    \caption{Kendall's $\tau$ scores for the datasets and results shown in Tables~\ref{tab:results:vision}~and~\ref{tab:results:time}.}
    \label{tab:kt}
    \begin{tabular}{@{}l*{5}{S[table-format=-1.1,table-model-setup=\bfseries,text-series-to-math,table-column-width=1.5cm]}@{}}
        \toprule
        \textbf{Dataset} & {\textbf{MeCo}} & {\textbf{NASWOT}} & {\textbf{SNIP}} & {\textbf{ZiCo}} & {\textbf{FLOPs}} \\
        \midrule
        CIFAR10 & 0.0 & 0.0 & -0.2 & \bfseries 0.6 & 0.0 \\
        CIFAR100 & 0.0 & 0.0 & -0.2 & \bfseries 0.6 & 0.0 \\
        ArxPhotos314 & 0.0 & 0.0 & \bfseries 0.2 & -0.2 & 0.0 \\
        Flowers & 0.2 & 0.2 & -0.4 & \bfseries 0.4 & 0.2 \\
        Birds & 0.0 & 0.0 & -0.2 & \bfseries 0.6 & 0.0 \\
        Pets & 0.1 & -0.4 & 0.2 & \bfseries 0.8 & -0.3 \\
        Cars & -0.2 & -0.2 & 0.0 & \bfseries 0.8 & -0.2 \\
        GTSRB & \bfseries 0.4 & -0.4 & 0.2 & -0.2 & -0.4 \\
        \midrule
        BitBrain & \bfseries 1.0 & 0.8 & -0.6 & -0.4 & -0.2 \\
        Mafaulda & -0.2 & 0.0 & \bfseries 0.6 & 0.4 & 0.6 \\
        Daliac & 0.1 & -0.1 & \bfseries 0.3 & 0.1 & -0.1 \\
        \bottomrule
    \end{tabular}
\end{table}

\subsection{Proxy Ensemble Analysis}
\label{subsec:evaluation:proxy}

We give a detailed analysis of the proxy \enquote{disagreement} we described in the previous section in Table~\ref{tab:kt}. To quantify \enquote{disagreement}, we compute the Kendall rank correlation coefficient (Kendall's $\tau$ score) to measure the ordinal association between each of the four proxy scores and the quantized post-training accuracy we presented in Tables.~\ref{tab:results:vision}~and~\ref{tab:results:time}. In addition, we provide the $\tau$ score between \glspl{FLOP} and accuracy.

The $\tau \in \left[-1, 1\right]$ score quantifies the strength and direction of monotonic relationships between two variables based on order, where $1$ describes a perfect agreement, i.e. the rankings are identical, $-1$ a perfect disagreement, i.e. the rankings are the opposite of each other, and $0$ denotes no correlation between the rankings of the two variables. Ideally, for zero-shot NAS proxies, a $\tau$ score close to $1$ is desirable, as this means that \glspl{DNN} with higher proxy scores will also achieve higher accuracy, while especially a score below $0$ is undesirable, as it misguides the \gls{NAS} algorithm during optimization.

\begin{figure}[t]
    \centering
    \subfloat[Image classification\label{fig:ktproxies:vision}]{%
        \includegraphics[width=0.34\textwidth]{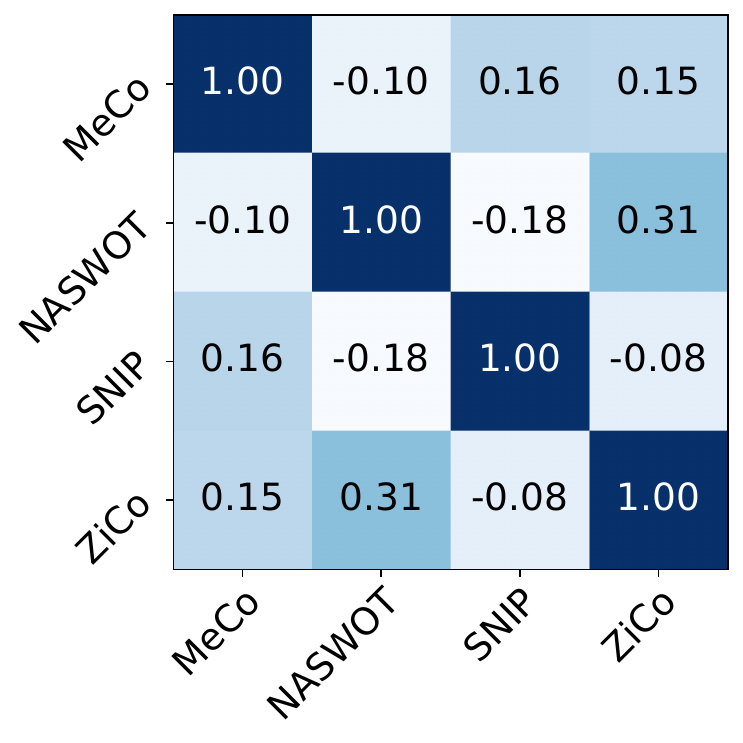}%
    }
    \subfloat[Time series classification\label{fig:ktproxies:time}]{%
        \includegraphics[width=0.34\textwidth]{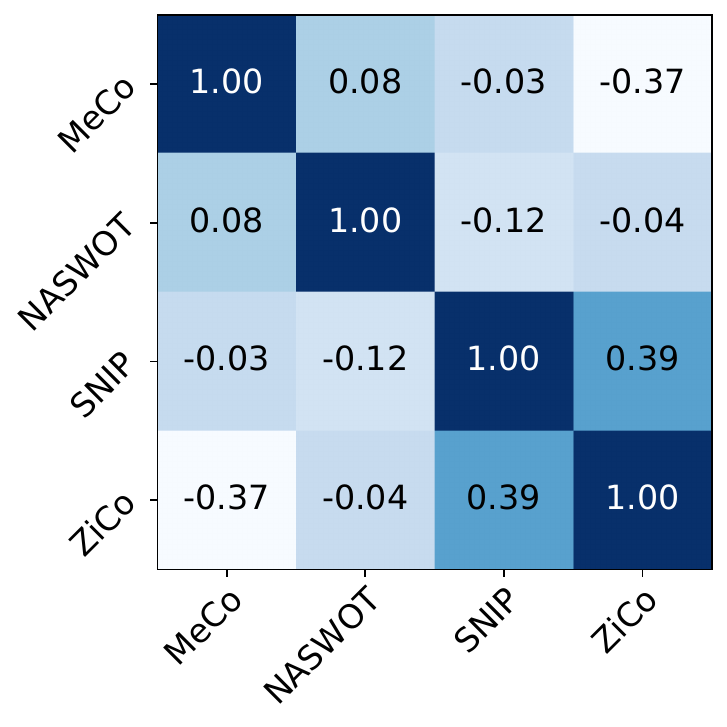}%
    }
    \caption{Kendall's $\tau$ scores between the four zero-shot proxies for the image and time series classification tasks.}
    \label{fig:ktproxies}
\end{figure}

In Table~\ref{tab:kt}, it can be seen that none of the four proxies has a consistent and high positive $\tau$ score for all to evaluate datasets. Even worse, the proxies sometimes have a negative relationship with accuracy for some datasets, while working well for others. This can be observed especially well with ZiCo and SNIP. On the other hand, it can be seen that for all datasets, at least one of the four proxies achieved a high positive $\tau$ score and, more importantly, outperformed the score achieved by \glspl{FLOP}. As a result, we concluded that for our search space (a) using a proxy ensemble is necessary to find a Pareto front of \glspl{DNN} that, when trained, achieve consistently good ranking across many datasets and prevent biases and inaccuracies of individual proxies from misguiding the exploration and (b) using zero-shot \gls{NAS} proxies is a better metric to rank models by approximated accuracy than using \glspl{FLOP}. 

Finally, in Fig.~\ref{fig:ktproxies}, we show the $\tau$ scores calculated among the proxies themselves for the image and time series optimization as heatmaps. Note that the lower and upper triangular matrices describe the same relationship. In Fig.~\ref{fig:ktproxies:vision} and~\ref{fig:ktproxies:time} we observe that for both optimizations the proxies generally did not have a strong relationship with each other, and in the few cases where they did, such as SNIP and ZiCo in Fig.~\ref{fig:ktproxies:time}, it was positive, i.e. the proxies were in agreement. These results confirm that the use of this specific ensemble of proxies provides PrototypeNAS with diverse and balanced evaluations, thus mitigating the biases and inaccuracies of individual proxies. Furthermore, our results are in agreement with~\cite{huang2025evolving}, who proposed the same ensemble of proxies.

\begin{figure}[t]
    \centering
    \includegraphics[width=.48\textwidth]{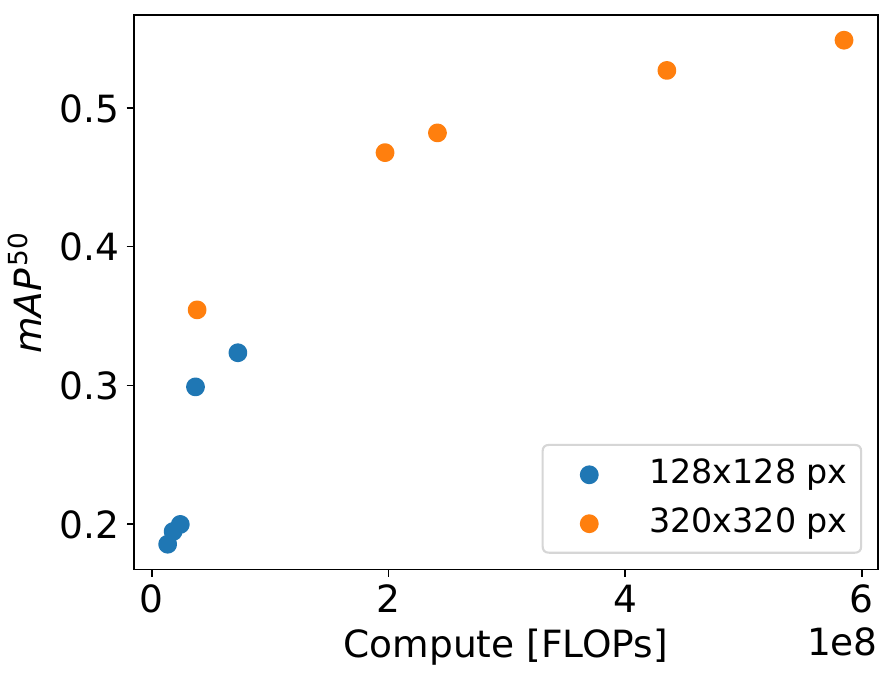}
    \caption{Tradeoff between $mAP^{50}$ and \glspl{FLOP} of the MBedNet YOLO architectures designed by PrototypeNAS.}
    \label{fig:detect:map}
\end{figure}

\subsection{Object Detection}
\label{subsec:evaluation:detection}

To further demonstrate the flexibility of PrototypeNAS, we applied it to an object detection task, specifically person detection, trained on the COCO dataset. We used the YOLOv5 framework, replacing its regular backbone network with MBedNet, which we also used in the results presented in Section~\ref{subsec:evaluation:classification}, but keeping the original YOLOv5 detector instead of a normal classification head. This allowed us to use the same search space for optimization as for the image classification task, while using the YOLOv5 data augmentation, loss function, anchor boxes, and training hyperparameters for training afterwards.

We ran two different experiments: The first one with an input image resolution of $128 \times 128$, using the same constraints as in Section~\ref{subsec:evaluation:classification}, and also targeting the iMXRT1062. The second one with a larger input image resolution of $320 \times 320$ and without any memory or \glspl{FLOP} constraints targeting a Raspberry Pi 5 SoC. Identical to the experiments in Section~\ref{subsec:evaluation:classification}, we used PrototypeNAS to explore 500 \glspl{DNN}, from which we selected a subset of $k=5$ architectures for training, pruning, quantization (\gls{PTQ}), and evaluation

We show the results between megaflops and $mAP^{50}$, the mean average accuracy of the model at 50\,\% intersection over union, a key performance indicator of object recognition models, for the five architectures selected after optimization and after 100 epochs of training for both the $128 \times 128$ and $320 \times 320$ pixel input resolutions (see Fig.~\ref{fig:detect:map}). The results show that PrototypeNAS was also able to work for use cases outside of the classification tasks that are usually the sole focus of zero-shot NAS research, providing a good $k=5$ selection, where each selected \gls{DNN} provides a meaningful tradeoff between the two evaluation metrics.

\subsection{Comparison with MCUNet and NATS-Bench}
\label{subsec:evaluation:comparison}

\begin{figure}[t]
    \centering
    \subfloat[TinyNAS (MCUNet in0-in4)\label{fig:compare:mcunet}]{%
        \includegraphics[width=0.48\textwidth]{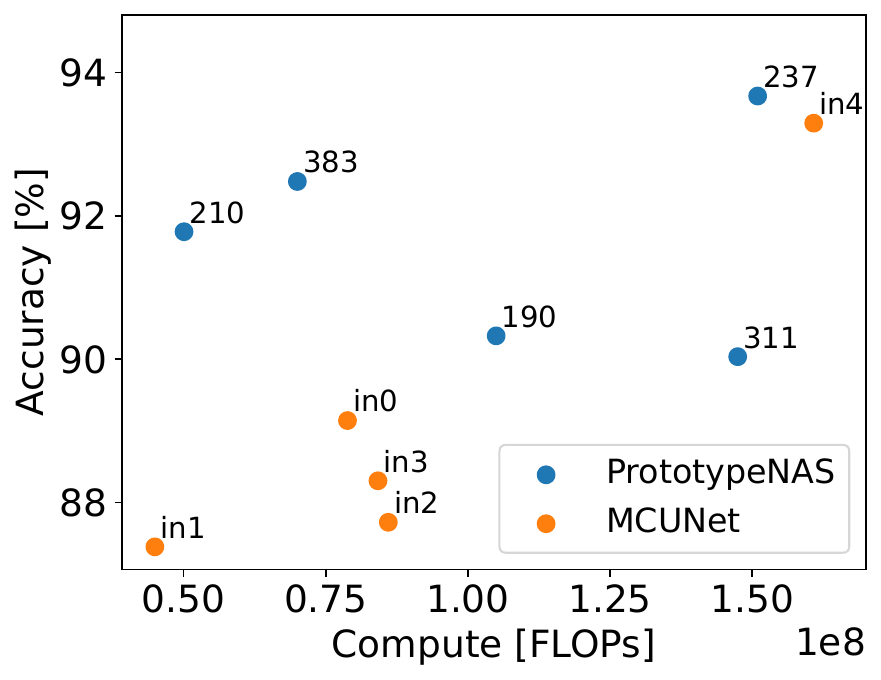}%
    }
    \hfill
    \subfloat[NATS-Bench\label{fig:compare:natsbench}]{%
        \includegraphics[width=0.48\textwidth]{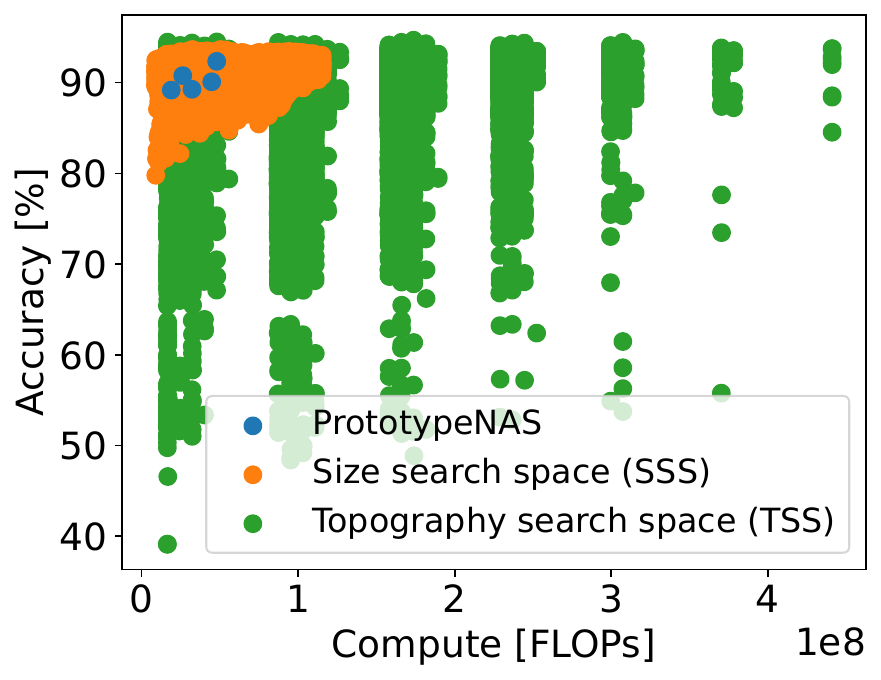}%
    }
    \caption{Comparison of PrototypeNAS with TinyNAS and NATS-Bench for CIFAR10.}
    \label{fig:compare}
\end{figure}

We compare PrototypeNAS to two related hardware aware \gls{NAS} methods on the CIFAR10 dataset, see Fig.~\ref{fig:compare}. First, a comparison with MCUNet which is a selection of five \glspl{DNN} searched with TinyNAS~\cite{lin2020mcunet}, and second a comparison with models from the two search spaces proposed in NATS-Bench~\cite{dong2021nats}.

\textbf{TinyNAS} is a two-step optimization method based on the Mobile search space~\cite{tan2019mnasnet} that is specifically designed for finding \glspl{DNN} for \gls{MCU} deployment. TinyNAS first optimizes the Mobile search search space to fit the resource constraints of an \gls{MCU} and then performs one-shot \gls{NAS}~\cite{bender2018understanding} only on the optimized subset to find an efficient model. The optimized search spaces are found by changing the input resolution and the model width hyperparameters. The quality of the sampled search spaces are evaluated by randomly sampling a number of \glspl{DNN} from them and evaluating the cumulative distribution function of the sampled \glspl{DNN}'s \glspl{FLOP}. In their paper, Lin et al.~\cite{lin2020mcunet} describe a set of five \glspl{DNN} (MCUNet in0-in4) which were designed using TinyEngine and for which the authors provide pre-trained weights on ImageNet.

We show a comparision between the \glspl{DNN} designed by PrototypeNAS with MCU\-Net for CIFAR10 in Fig.~\ref{fig:compare:mcunet}. For PrototypeNAS, we used the same results shown in Table~\ref{tab:results:vision}, while for MCUNet we fine-tuned the pre-trained ImageNet weights for 100 epochs on CIFAR10. The results show that the five \gls{DNN} proposed by PrototypeNAS consistently outperform MCUNet in terms of accuracy, sometimes by as much as 5\,\%, at roughly the same computational cost. This shows that compared to other zero-shot search spaces that limit their search to finding subnetworks within a single large super-network architecture, expanding the search space to include pruning and different architecture classes, as is done in PrototypeNAS, allows finding models with higher knowledge density within the same resource footprint.

\textbf{NATS-Bench} is a \gls{NAS} benchmark for optimizing both \gls{DNN} topology and size. NATS-Bench contains two search spaces, one with $15,625$ trained \gls{DNN} candidates focusing on architecture topology (TSS) and one with $32,768$ \gls{DNN} candidates focusing on architecture size (SSS). NATS-Bench uses a cell-based approach where the skeleton of each cell is constructed from remaining blocks with a fixed topology. In the size search space, the number of channels of each cell is optimized, effectively allowing the \gls{DNN} candidates to scale in width. The topology search space, on the other hand, optimizes a predefined set of operations within each cell, i.e., the combination of layers.

The comparison between PrototypeNAS and the two NATS-Bench search spaces is shown in Fig.~\ref{fig:compare:natsbench}. For NATS-Bench, we plot all \glspl{DNN} that have an accuracy higher than 70\,\% on CIFAR10 after the maximum number of training epochs reported by the authors. For PrototypeNAS, we again use the results shown in Table~\ref{tab:results:vision} for CIFAR10, with the only difference that we re-trained the models for 100 epochs on a smaller input image resolution of $69 \times 69$ pixels instead of $128 \times 128$ to more closely match the $32 \times 32$ pixel resolution used by NATS-Bench. We chose $69 \times 69$ pixels because it was the smallest input size supported by all the \glspl{DNN}. The results show that the \glspl{DNN} architectures designed by PrototypeNAS are again competitive with the best architectures found by NATS-Bench in both of its search spaces. However, it should be noted that for the NATS-Bench results, each of the $48,393$ \glspl{DNN} had to be trained separately for up to 200 epochs, while for PrototypeNAS only 5 \glspl{DNN} had to be trained for 100 epochs to obtain models with similar results.

\subsection{Impact on Performance and Emissions}
\label{subsec:evaluation:ablation}

\begin{figure}[t]
    \centering
    \subfloat[Accuracy versus FLOPs\label{fig:ablation:accflops}]{%
        \includegraphics[width=0.48\textwidth]{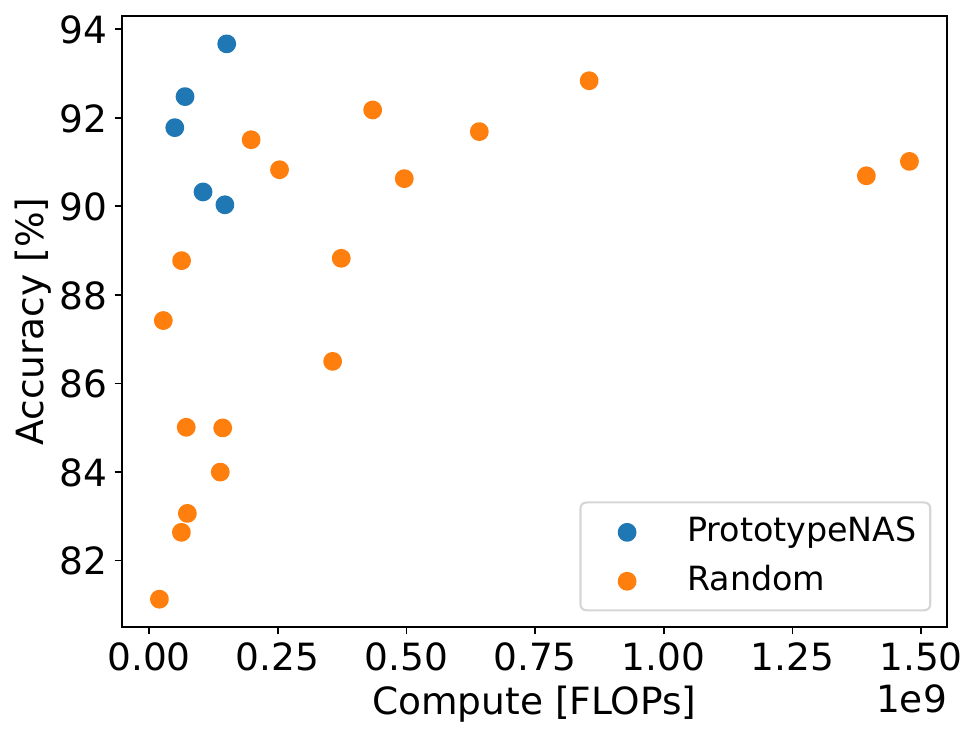}%
    }
    \hfill
    \subfloat[Duration and emission analysis\label{fig:ablation:emission}]{%
        \includegraphics[width=0.48\textwidth]{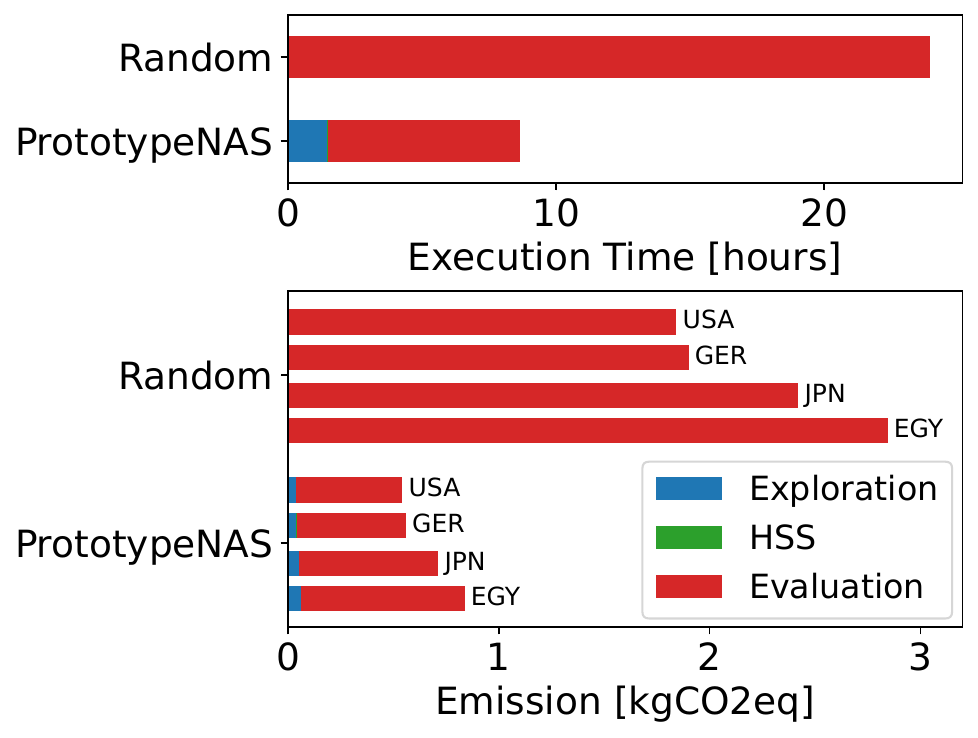}%
    }
    \caption{Performance and emission comparison of PrototypeNAS and randomly selected models on CIFAR10.}
    \label{fig:ablation}
\end{figure}

To better assess the impact of PrototypeNAS's unique combination of \gls{HSS}, Bayesian optimization, and zero-shot \gls{NAS} on the performance of designed \glspl{DNN}, we disabled the three components and repeated the CIFAR10 experiment from Section~\ref{subsec:evaluation:classification}. In this experiment, besides \gls{DNN} accuracy and FLOPs, we also focus on evaluating the execution time and estimated emissions of the \gls{NAS}. We do this, because a major motivation for introducing \gls{HSS}, Bayesian optimization, and zero-shot \gls{NAS} into PrototypeNAS was to increase the method's sample efficiency and to reduce its search time and environmental footprint.

After disabling \gls{HSS}, Bayesian optimization, and zero-shot \gls{NAS}, we randomly sampled 20 model from the search space in Sec.~\ref{subsec:method:exploration}. Then, we filtered all models that were infeasible according to the constraints we described in Section~\ref{subsec:evaluation:classification}. All reminaing models were pretraining on ImageNet and then trained and quantized on CIFAR10 with the same hyperparameters and training configuration as in Section~\ref{subsec:evaluation:classification}. In Fig.~\ref{fig:ablation}, we compare the CIFAR10 results from Tab.~\ref{tab:results:vision} that we achieved with the full PrototypeNAS algorithm to the random sampled resuls.

Fig.~\ref{fig:ablation:accflops} visualizes the trade-offs between quantized accuracy and FLOPs for the PrototypeNAS models (blue) and random models (orange) on CIFAR10. Even when sampled randomly, it can be seen that the search space we propose in Sec.~\ref{subsec:method:exploration} yields models that, while dominated by the models found by the full PrototypeNAS algorithm, still have a high accuracy when trained and quantized (often only 2-3\,\% behind in accuracy).

We used CodeCarbon~\cite{benoitcourty2026codecarbon} to assess the execution time and approximate emissions of the random and PrototypeNAS models on a local desktop for four different regions (USA, Germany, Japan, and Egypt), see Fig.~\ref{fig:ablation:emission}. The desktop has 16 GB RAM, an eight-core AMD Ryzen 7 2700 processor, and a GeForce RTX 2070 GPU. Despite the optimization and \gls{HSS} overhead introduced by the PrototypeNAS algorithm, the benefit of training only five models instead of 17 feasible random models still resulted in a 62\,\% reduction in execution time, which is reflected in a similar average decrease in emissions across the four analyzed regions. However, when comparing the average emissions of the regions amongst each other, a noticeable difference can be observed, highlighting the impact of global and structural factors on the sustainability of running \gls{NAS}.
\section{Conclusion}
\label{ref:conclusion}

We propose PrototypeNAS, a novel three-step zero-shot \gls{NAS} method for rapidly designing \glspl{DNN} for \glspl{MCU}. Unlike previous work, we perform \gls{MOO} using an ensemble of zero-shot proxies to optimize over many baseline architectures, rather than just sampling subsets from a single one. This allows us to find \gls{DNN} architectures within minutes that achieve competitive accuracies on 12 different datasets for tasks such as image classification, time series classification, and object detection. The found \glspl{DNN} run on standard \glspl{MCU} and outperform architectures found using other hardware-aware \gls{NAS} methods such as TinyNAS and NATS-Bench on CIFAR10.

\begin{credits}
\subsubsection{\ackname}
This work was funded by the European Commission as part of the MANOLO project under the Horizon Europe programme Grant Agreement No.101135782.
\subsubsection{\discintname} The authors have no competing interests to declare that are relevant to the content of this article.
\end{credits}

\bibliographystyle{splncs04}
\bibliography{references}

\end{document}